\documentclass{article}

\usepackage{microtype}
\usepackage{graphicx}
\usepackage{subcaption}
\usepackage{booktabs}
\usepackage{hyperref}

\usepackage[accepted]{icml2026}
\makeatletter
\renewcommand{\ICML@appearing}{\textit{ICML 2026 Workshop on Continual Adaptation at Scale (CATS)}, Seoul, South Korea. Copyright 2026 by the author(s).}
\makeatother

\usepackage{amsmath}
\usepackage{amssymb}
\usepackage{mathtools}
\usepackage{amsthm}
\usepackage[capitalize,noabbrev]{cleveref}
\usepackage{algorithm}
\usepackage{algorithmic}
\usepackage{xcolor}
\usepackage{multirow}

\icmltitlerunning{Audit Before You Merge: Provenance, Probing, and Continual LoRA Composition}

\begin{document}

\twocolumn[
\icmltitle{Audit Before You Merge: Provenance, Probing, \\
           and Continual LoRA Composition}

\icmlsetsymbol{equal}{*}
\begin{icmlauthorlist}
% \icmlauthor{Anonymous Authors}{anon}
\icmlauthor{Lenny Aharon}{col}
\icmlauthor{Neta Glazer}{bi}
\icmlauthor{Lior Aharon}{mit}
\end{icmlauthorlist}
% \icmlaffiliation{anon}{Anonymous Institution}
\icmlaffiliation{col}{Columbia University, New York, NY, USA}
\icmlaffiliation{mit}{MIT, Cambridge, MA, USA}
\icmlaffiliation{bi}{Bar-Ilan University, Ramat Gan, Israel}
\icmlcorrespondingauthor{Lenny Aharon}{lenny.aharon@columbia.edu}
\icmlcorrespondingauthor{Neta Glazer}{neta.glazer@biu.ac.il}
\icmlcorrespondingauthor{Lior Aharon}{liorah27@mit.edu}
\icmlkeywords{LoRA, PEFT, model merging, continual adaptation, sustainable AI, evaluation, adapter audit}
\vskip 0.3in
]

\printAffiliationsAndNotice{}

\begin{abstract}
% Post-hoc composition of pre-trained LoRA adapters is an attractive,
% gradient-free route to continual adaptation: base weights stay frozen
% and PEFT modules can be sourced from public hubs. We make this
% composition more reliable through two infrastructure contributions:
% (i)~a two-check adapter provenance audit (base-model match plus
% non-trivial delta norm) that catches defective public adapters whose
% inclusion inflates measured interference by ${\sim}10$pp; and
% (ii)~a leakage-free probing protocol with disjoint splits and
% single-adapter reference evaluations that decouple adapter quality
% from merge quality. On an audited 4-adapter Llama-3.1-8B-Instruct
% set, naive merging causes $-10.5$pp on GSM8K and $-18.3$pp on
% HumanEval, while uniform scaling (TA $\lambda{=}0.5$) sits within a
% standard error of the unmerged baseline. A coordinate search over
% per-adapter coefficients including \emph{negative} values shows the
% math adapter is probe-optimal at $\lambda{=}-0.5$, outside the
% search space of positive-only mergers. The same audit and merge
% protocol applied to Mistral-7B-Instruct-v0.3 catches 2/4 candidate
% adapters and reproduces the TA-near-baseline / naive-drops /
% DARE-worst pattern on GSM8K, supporting that our findings are not
% Llama-specific.
Post-hoc merging of public LoRA adapters is an attractive
gradient-free path to continual adaptation, but reliable measurement
of catastrophic forgetting in this setting depends on infrastructure
that prior work has not addressed. We treat each adapter merge as a
sequential task in a continual-learning setting, with the
instruction-tuned base model's capabilities as Task~0, and study
forgetting on Llama-3.1-8B-Instruct. Public adapter hubs contain
modules that fail basic provenance checks: silent base-version
mismatches and near-zero delta norms whose inclusion inflates
measured interference by ${\sim}10$pp. We propose a two-check audit
(base-model match, non-trivial delta norm) and a leakage-free
probing protocol as preconditions for any forgetting study in this
setting. On an audited 4-adapter set, naive merging causes $-10.5$pp
on GSM8K and $-18.3$pp on HumanEval, while Task Arithmetic at
$\lambda{=}0.5$ sits within a standard error of the unmerged
baseline. A coordinate search including \emph{negative} coefficients
identifies the math adapter as probe-optimal at $\lambda{=}-0.5$,
outside the reach of positive-only mergers, suggesting forgetting
has structure that negation-aware merging may exploit. The same
protocol replicates on Mistral-7B-Instruct-v0.3, where 2 of 4
candidate adapters fail the audit and the audited merge reproduces
the same merge-method ranking on GSM8K, supporting that our findings
are not Llama-specific.
\end{abstract}

%============================================================
\section{Introduction}
%============================================================

Continual adaptation of large language models without retraining
their base weights is increasingly attractive: capabilities
accumulate across tasks, foundation models stay frozen, and the
parameter-efficient fine-tunes (PEFTs) needed for new tasks are
available on public hubs. Post-hoc composition of pre-trained
Low-Rank Adaptation (LoRA) adapters~\citep{hu2022lora} is one
realization. Each new adapter's delta is folded into the model's
weights once, before inference, without further training, original
training data, or gradients. We treat this as a continual-learning
problem: each adapter merge is a sequential task, the
instruction-tuned base model's capabilities (math, code, instruction
following) constitute Task~0, and the central question is how much
these prior capabilities degrade as further tasks are added.
Proposed methods include Task Arithmetic~\citep{ilharco2022editing},
TIES~\citep{yadav2023ties}, DARE~\citep{yu2024dare},
MagMax~\citep{marczak2024magmax}, and more recently
PICO~\citep{crowded_bspace2026} and LoRM~\citep{closedform_peft2024}.
Published evaluations regularly report catastrophic forgetting
figures in the 20+ pp range against naive merging, closed by these
methods.

Reliable measurement of forgetting in this setting turns out to
require infrastructure that prior work has not addressed. In
preparing this study we found that public LoRA hubs contain adapters
that fail basic provenance checks. One Llama-3.1 ``code'' adapter
has total delta norm $0.02$, so every forward pass with it applied
is within floating-point noise of the base model. Another, listed
for Llama-3.1 use, was trained on \emph{Llama-3} (not 3.1) and
produces a $\sim$46-norm delta when applied to the wrong base,
yielding visible-but-meaningless behavioral change. \emph{Including
such adapters in a sequential merge can generate the appearance of
catastrophic forgetting from a defective constituent.} On a random
sample of 80 public LoRAs, only $7.5\%$ pass our audit for their
declared base ($66\%$ of loadable adapters have a base-model
mismatch; App.~\ref{app:prevalence}). A controlled ablation
(App.~\ref{app:ablation}) attributes essentially all of the resulting
$12$pp swing in full-test GSM8K naive-merge damage
($-21.7\text{pp}\!\to\!-9.2$pp under audit) to the single wrong-base
adapter; the near-zero-norm adapter is a no-op, as predicted.

Our position is therefore that progress on catastrophic forgetting
in post-hoc LoRA merging is bottlenecked less on new
merging methods than on the measurement infrastructure needed to
evaluate them. Concretely, we contribute:

\begin{enumerate}\itemsep1pt
\item \textbf{A clean-adapter characterisation of forgetting.}
  On audited adapters, naive merging causes $-10.5$pp on GSM8K and
  $-18.3$pp on HumanEval; Task Arithmetic at $\lambda{=}0.5$ is
  within stderr of the unmerged baseline. The near-zero gap for Task
  Arithmetic partly reflects our adapter set, since no audited
  adapter helps GSM8K beyond one-sigma noise in isolation and the
  unmerged baseline is a competitive ceiling
  (App.~\ref{app:upperbounds}, Sec.~\ref{sec:results}).
\item \textbf{Two evaluation preconditions missing from prior work.}
  A two-check adapter provenance audit (base-model match, non-trivial
  delta norm), and a leakage-free probing protocol with disjoint
  probe/$\alpha$-selection/test splits and single-adapter reference
  evaluations that decouple adapter quality from merge quality
  (Sec.~\ref{sec:setup};
  App.~\ref{app:provenance},~\ref{app:upperbounds},~\ref{app:probing}).
\item \textbf{Preliminary evidence for negation-aware merging.}
  A gradient-free per-adapter coefficient search with \emph{negative}
  candidates identifies $\lambda_{\text{math}}{=}-0.5$ as optimal for GSM8K, outside positive-only mergers' search
  space. On full tests it trades math plasticity for protected-task
  retention rather than beating Task Arithmetic or TIES, but
  suggests forgetting has structure that future negation-aware
  methods may exploit (App.~\ref{app:negation}).
\item \textbf{Cross-family replication on the protected task.}
  On Mistral-7B-Instruct-v0.3 the audit catches 2 of 4 candidate
  adapters on silent base-version mismatch, and the audited
  2-adapter Mistral merge reproduces the Llama pattern on GSM8K:
  Task Arithmetic near baseline, naive merging drops, DARE
  worst~(Table~\ref{tab:mistral-merging}).\footnote{HumanEval
  cross-family is deferred due to a known harness-template
  sensitivity for Mistral instruct.} Thus, our protected-task findings are not Llama-specific.
\end{enumerate}

%============================================================
\section{Related Work}
\label{sec:related}
%============================================================
\paragraph{Post-hoc merging.} Magnitude-based mergers modulate
per-parameter weights: Task Arithmetic
(TA)~\citep{ilharco2022editing} scales each delta by a scalar
$\lambda$; TIES~\citep{yadav2023ties} trims small magnitudes and
resolves sign conflicts; DARE~\citep{yu2024dare} drops random
parameters and rescales; MagMax~\citep{marczak2024magmax} keeps the
larger-magnitude element per coordinate. Structure-aware mergers
exploit delta algebra: PICO~\citep{crowded_bspace2026} damps
adapter-shared SVD directions; LoRM~\citep{closedform_peft2024}
gives a closed-form PEFT solution; conflict-aware~\citep{cofima2024}
and gating~\citep{hoy2025stable} extensions also exist.
Surveys~\citep{yang2026model, yadav2024moerging, shenaj2025k} and an
in-the-wild study~\citep{hititsystematic} find uniform
scaling most reliable across LLM merges. We do not propose a new
merger; we re-evaluate the existing family on \emph{audited}
adapters and report that TA and TIES approach the protected-task
retention implied by single-adapter references, leaving little
headroom in this setting.

\paragraph{LoRA-specific composition.} A line of work argues
uniform scaling underexploits LoRA's low-rank structure.
SA-LoRA~\citep{li2025salora} aligns adapters before merging;
task-singular-vector sharing~\citep{gargiulo2025tsv} composes via
shared SVD bases; intra-layer-significance~\citep{ila2024} and
merging-with-safety~\citep{led_merging2025} explore richer
per-direction handling. A more recent line of LoRA-aware mergers
includes LoRA-LEGO~\citep{lora_lego2024} (rank-wise clustering for
modular composition), OSRM~\citep{osrm2025} (orthogonal subspaces
to reduce interference), TARA~\citep{tara2026} (preference-aligned
merging preserving subspace coverage), and LoRI~\citep{lori2025}
(frozen random projection matrices with task-specific sparsity
masks). We add a controlled orthogonal-projection merger
representative of this family (App.~\ref{app:orthogonal}): the audit's
effect on it ($+11.1$pp GSM8K) matches its effect on naive merging
within $1$pp and dwarfs the merger's $+2$pp edge over naive on the
audited pool, so structure-aware merging is complementary to, not a
substitute for, auditing. Faithful reproductions of SA-LoRA, OSRM,
LoRA-LEGO and LoRI under the audit are future work. Our negation
finding (App.~\ref{app:negation}) supports the broader thread: even
scalar-per-adapter coefficients, allowed to take negative values,
expose behavior outside positive scaling.

\paragraph{Continual learning and continual merging.} We operate post-hoc on
already-trained adapters, distinct from training-time continual
learning~\citep{mccloskey1989catastrophic, kirkpatrick2017overcoming,
zenke2017continual,glazer2025few} but inheriting its central concern: how prior
tasks degrade as new ones are added. Post-hoc merging is the
rehearsal-free, gradient-free corner of this taxonomy; scalable
batch~\citep{scalable_continual_merging2025} and on-device
online~\citep{shenaj2025k} variants, surveyed
in~\citep{shi2025continual}, cover the surrounding landscape. Our
block-importance probe is the data-free analogue of Fisher
importance.

\paragraph{The missing audit step.}
\citet{hammoud2024model} shows a single poorly-trained constituent
destabilises a merge, and reproducibility-gap
studies~\citep{hititsystematic} report cross-paper
inconsistency, but none verify adapters against their declared base
or delta norm before use.
\citet{liu2026appeal} recycle ${\sim}$1{,}000 Hub LoRAs for
Llama-3.1-8B-Instruct but filter only shape mismatches and NaN
weights, which cannot detect base-version mismatch.
\emph{To our knowledge no
prior work systematically audits public LoRAs before merging.} Our protocols (leakage-free probing,
single-adapter references) also address gaps in evaluation
practice: hyperparameter selection and final evaluation share data
in many published pipelines, and merge scores are reported without
single-adapter references that would distinguish merge damage from
baseline adapter-quality artefacts. In our setting, applying the
audit changes the full-test GSM8K naive-merge reading by $12.4$pp
($-21.7\text{pp}\to-9.2\text{pp}$; App.~\ref{app:ablation}), a
swing that would suffice to flip qualitative conclusions about which
mergers are ``catastrophic'' versus ``acceptable''. Adapter-quality controls
deserve to be a precondition for merging benchmarks, not an
afterthought.

%============================================================
\section{Experimental Setup and Adapter Audit}
\label{sec:setup}
%============================================================

\paragraph{Base model.} Llama-3.1-8B-Instruct (32 transformer blocks,
224 LoRA-target layers).

\paragraph{Adapter audit.} We apply two provenance checks before
admitting any public LoRA into the merge set:

\begin{itemize}\itemsep1pt
\item \textbf{Base-model match.} The adapter's declared base must
  exactly equal the base model under test (e.g., Llama-3.1-8B-Instruct,
  not Llama-3-8B).
\item \textbf{Non-trivial delta norm.} Total Frobenius norm of the
  reconstructed delta $\big(\sum_\ell \|B_\ell A_\ell s\|_F^2\big)^{1/2}$
  (with $s = \alpha/r$ or $\alpha/\sqrt{r}$ per the config) must exceed $1.0$. We threshold the reconstructed weight modification rather than $(A, B, \alpha, r)$ directly because it is invariant to factorization choices and determines the behavioral change at inference. Audit decisions are insensitive to threshold choice in $[0.1, 1.5]$ on our candidate set: the defective adapter has norm $0.02$, roughly $50$--$800\times$ below the four audited adapters (norms $1.07$--$16.62$); see App.~\ref{app:audit-ablation}.
\end{itemize}

Applied to candidate public LoRAs for Llama-3.1-8B-Instruct, the
audit flags two adapters: one with total delta norm $0.02$ (a no-op
in practice) and one trained on Llama-3, not Llama-3.1, whose
mismatched delta accounts for the swing in measured
interference reported in Sec.~\ref{sec:related}. Full failure-mode
descriptions are in App.~\ref{app:provenance}.

\paragraph{Audited 4-adapter set.} Four clean, audited Llama-3.1
LoRAs spanning distinct capabilities: \emph{math} (total delta norm
$16.6$), \emph{code} ($1.7$), \emph{general\_nlp} ($1.1$), and
\emph{medical} ($1.6$). The asymmetric norms reflect real-world
deployment.

\paragraph{Cross-family check.} The same audit applied to a
candidate Mistral-7B-Instruct-v0.3 4 adapter set flags 2 of 4
adapters on a minor-version mismatch (declared v0.1 / v0.2 against
v0.3). Both failing adapters pass a norm-only check, so the
base-version check is what catches them; merge results on the
audited Mistral pair are in Sec.~\ref{sec:results}.

\begin{figure}[t]
\centering
\includegraphics[width=\linewidth]{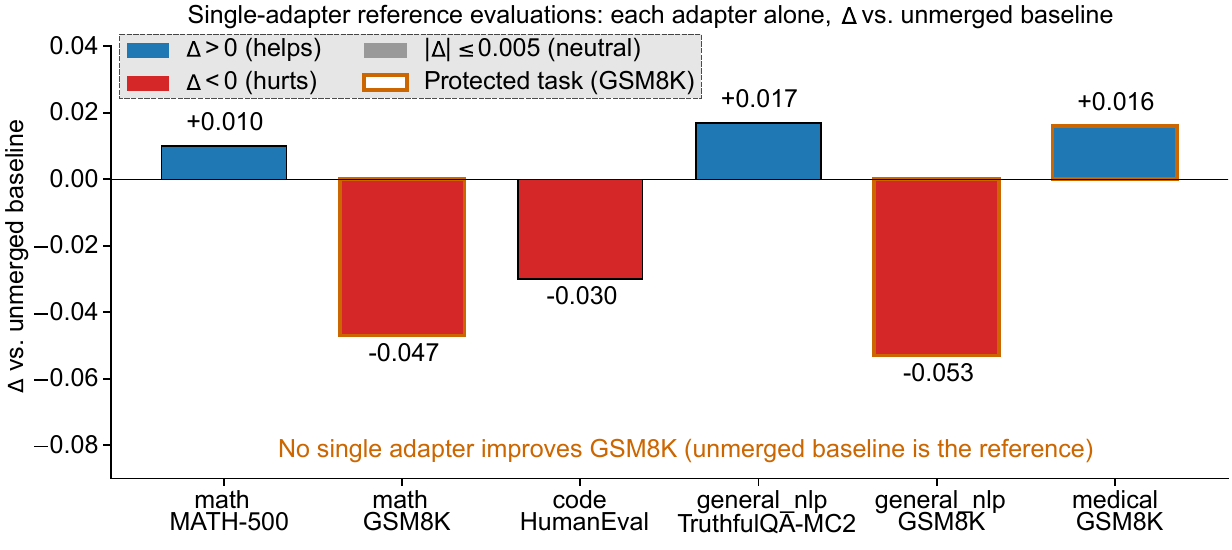}
\caption{Single-adapter reference evaluations. For each (adapter, benchmark) pair, the bar shows the signed change vs.\ the unmerged baseline when only that adapter is applied (no merging, no other adapters). GSM8K bars (the protected task) have orange outlines.}
\label{fig:upperbounds}
\end{figure}

\paragraph{Benchmarks.} GSM8K (1{,}319 problems, strict-match
exact-answer scoring); HumanEval (164 problems, pass@1); MATH-500
(500 problems, 4-shot~\citep{hendrycks2021math}). Decoding
parameters, seeds, and compute environment are in
App.~\ref{app:repro}.

\paragraph{Leakage-free probing.} We designate GSM8K as the
\emph{protected task} (every audited adapter measurably shifts it, and strict-match scoring is robust to decoding artifacts; HumanEval
and MATH-500 are reported alongside as non-protected diagnostics)
and partition it into three disjoint subsets: a \textbf{probe set}
($n{=}50$, train examples), an \textbf{$\alpha$-selection set}
($n{=}200$, train indices 50--250), and a \textbf{test set} (full
$1{,}319$, used only for final reporting). Full protocol in
App.~\ref{app:probing}.

\begin{figure*}[t]
\centering
\includegraphics[width=\textwidth]{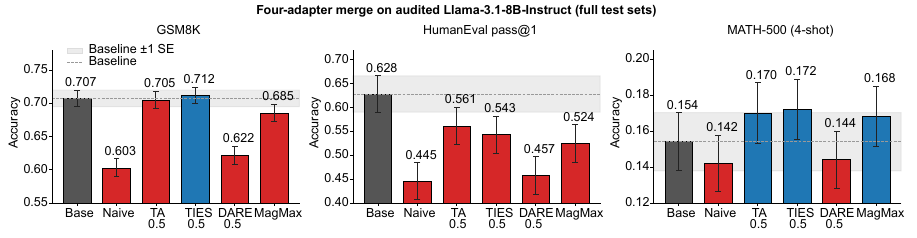}
\caption{Per-benchmark comparison of merge methods on the four
audited Llama-3.1-8B-Instruct adapters. Dashed line: unmerged
baseline. TA $\lambda{=}0.5$ and TIES $d{=}0.5$ sit within a
standard error of baseline on GSM8K and preserve HumanEval best
among merging methods; naive merge and DARE lose substantial
HumanEval and GSM8K. Fig.~\ref{fig:negation} shows where residual
headroom may still exist.}
\label{fig:methods}
\end{figure*}

\paragraph{Single-adapter reference evaluations.} We evaluate each adapter alone, decoupling its isolated capability from merge damage
(Fig.~\ref{fig:upperbounds}; full numbers in
App.~\ref{app:upperbounds}). Of the six (adapter, benchmark) pairs tested, only one shows a clearly positive change (math on MATH-500,
$+1$pp); general\_nlp lifts TruthfulQA-MC2 by $+2.6$pp (${\sim}1.7\sigma$,
borderline); medical/GSM8K is within noise; the remaining three hurt their respective benchmarks. Crucially, no single adapter helps the protected task GSM8K in isolation. This does not prove that no combination can help GSM8K, but it makes the unmerged baseline the relevant reference point for protected-task retention and explains
why methods that preserve the unmerged baseline are already strong on this audited set.

\paragraph{Merge methods.} We compare naive sum, Task Arithmetic ($\lambda{=}0.5$), TIES ($d{=}0.5$), DARE ($d{=}0.5$), and MagMax.
We merge the four deltas into one weight update,
$\theta_0 + \sum_t f(\Delta_t)$, applied before inference with no
inter-task retraining. Hyperparameters follow the original papers' recommended values; App.~\ref{app:fullresults} reports a
sensitivity sweep.
% All four deltas are combined into a single weight modification $\theta_0 + \sum_t f(\Delta_t)$, applied once before inference and
% not retrained between tasks. Hyperparameters use the recommended
% values from each method's original paper rather than being re-tuned
% on our adapter set; we report a sensitivity sweep in
% App.~\ref{app:fullresults}.

%============================================================
\section{Results}
\label{sec:results}
%============================================================

\textbf{Interference is substantial on clean adapters.}
Table~\ref{tab:main} and Fig.~\ref{fig:methods} report the full
comparison on the audited 4-adapter Llama-3.1-8B-Instruct set.
Naive sequential merging drops GSM8K from $0.707$ to $0.603$
($-10.5$pp) and HumanEval from $0.628$ to $0.445$ ($-18.3$pp),
while MATH-500 is essentially unchanged ($0.154\to 0.142$). The
asymmetry across benchmarks indicates task-specific rather than
uniform degradation. These drops are notably smaller than the
$20$pp+ figures reported in some prior work, consistent with our
claim in Sec.~\ref{sec:related} that unaudited adapters inflate
measured interference: a 4-adapter merge that includes the
defective Llama-3 adapter shifts naive-merge GSM8K damage from
$-10.5$pp to $-21.7$pp (controlled ablation in
App.~\ref{app:ablation}).

\begin{table}[!htbp]
\centering
\small
\caption{Llama-3.1-8B-Instruct, sequential merge of four audited
adapters. Scores on full test sets. Bold: best non-baseline per
column. Standard errors and additional ablations in
App.~\ref{app:fullresults}.
A current-harness reproduction of the naive row gives $0.615 / 0.476 / 0.142$
(${\sim}{+}1$pp drift, within stderr); the controlled ablation in
App.~\ref{app:ablation} uses the reproduction values as its anchor.}
\label{tab:main}
\begin{tabular}{lccc}
\toprule
Method & GSM8K & HE\,p@1 & M-500 \\
\midrule
Baseline (no merge)         & 0.707 & 0.628 & 0.154 \\
\midrule
Naive merge                 & 0.603 & 0.445 & 0.142 \\
Task Arith.~$\lambda{=}0.5$ & 0.705 & \textbf{0.561} & 0.170 \\
TIES $d{=}0.5$              & \textbf{0.712} & 0.543 & \textbf{0.172} \\
DARE $d{=}0.5$              & 0.622 & 0.457 & 0.144 \\
MagMax                      & 0.685 & 0.524 & 0.168 \\
\bottomrule
\end{tabular}
\end{table}

\textbf{Simple uniform scaling is near-optimal.} Task Arithmetic at
$\lambda{=}0.5$ recovers GSM8K to $0.705$ (within $0.2$pp of baseline)
and HumanEval to $0.561$; TIES at $d{=}0.5$ is comparable and
nominally exceeds baseline on GSM8K ($0.712$, within one stderr).
MagMax falls between TA/TIES and naive/DARE; DARE is close to naive.
The ordering is robust to a $d{=}0.3$ sweep (App.~\ref{app:fullresults}).

\textbf{Negation uncovers a diagnostic regime.} A per-adapter
coordinate search makes the math adapter probe-optimal at
$\lambda{=}-0.5$; the resulting configuration gains $+8.5$pp on the
GSM8K probe (Fig.~\ref{fig:negation}), in a region positive-only
mergers cannot reach (App.~\ref{app:negation}).

%============================================================
\section{Discussion and Limitations}
%============================================================
The audit-then-merge protocol yields three takeaways. First, adapter
audits matter: defective adapters inflate reported interference by
${\sim}10$pp -- an adapter-quality artifact, not merge behavior.
Second, on audited adapters, uniform scaling (TA $\lambda{=}0.5$, TIES
$d{=}0.5$) sits within a standard error of the unmerged GSM8K
baseline; the near-zero gap partly reflects that no audited adapter
helps the protected task alone (App.~\ref{app:upperbounds}), so prior
large method-level advantages over TA may reflect
pathological-adapter damage rather than a general property of the
problem. Third, the pattern replicates on Mistral-7B-Instruct-v0.3
(audit catches 2 of 4; TA $>$ TIES $>$ naive $>$ DARE on GSM8K;
Table~\ref{tab:mistral-merging}), with one informative divergence: the
math adapter helps the protected task on Mistral ($+2.9$pp) but not Llama, so the merge pattern holds even when single-adapter behavior
shifts.

Negation deserves separate treatment: the coordinate search makes the
math adapter probe-optimal at $\lambda{=}-0.5$, a setting no
scalar-positive merger can express. On full tests it beats naive on
every benchmark but not TA/TIES, so we present it as a search-space
finding, not a recommended merger; a negation-aware merger that
preserves plasticity on the negated adapter's own task is open
(App.~\ref{app:negation}).

\textbf{Robustness.} The conclusions are not GSM8K-specific: with
HumanEval as the protected task the same ordering holds (naive
$-18.3$pp; TA $-6.7$; TIES $-8.5$; Table~\ref{tab:main}), and the
naive-drops / uniform-scaling-recovers pattern persists across seeds and temperature-sampled decoding
(App.~\ref{app:robustness}). Per-step forgetting curves
(App.~\ref{app:perstep}) and a cross-scale prevalence audit spanning
1B--8B and Qwen (App.~\ref{app:prevalence}) corroborate.

\textbf{Limitations.} The prevalence audit spans four families across
three scales (App.~\ref{app:prevalence}), but the \emph{merge}
experiments are on two $7$--$8$B families; cross-scale merging is
future work. Provenance-and-norm checks (supplemented by content-based
signals, App.~\ref{app:augmented}) are necessary but not sufficient:
an adapter can pass yet be merely weak (hence the single-adapter
references, App.~\ref{app:upperbounds}) or be high-norm yet
semantically harmful (backdoors, jailbreaks), an orthogonal safety
problem. The negation search is a 4-adapter $\times$ 4-candidate
sweep on an $n{=}200$ probe; we report it as a search-space finding,
not a method. TA, TIES, and DARE use recommended defaults
(App.~\ref{app:fullresults}).

%============================================================
\bibliography{references}
\bibliographystyle{icml2026}

\newpage
\appendix
\onecolumn

%============================================================
\noindent\emph{Code:} \url{https://github.com/LennyAharon/audit_continual_learning}

\section{Adapter Provenance Details}
\label{app:provenance}
%============================================================

\textbf{Audited adapter set (used in this study).} Table~\ref{tab:provenance}
lists the four adapters after applying the audit. All four pass the
base-model match and non-trivial delta norm checks.

\begin{table}[h]
\centering
\small
\caption{Provenance of the four audited adapters. ``Norm'' is the
total Frobenius norm of the reconstructed delta, aggregated over all
LoRA-target layers ($(\sum_\ell \|\Delta_\ell\|_F^2)^{1/2}$).
``Single-adapter score'' is the test-set score
when only that adapter is applied to the primary target benchmark
(from App.~\ref{app:upperbounds}). HuggingFace identifiers and
checkpoint fingerprints for every candidate adapter are in
Table~\ref{tab:adapter-fingerprints}.}
\label{tab:provenance}
\begin{tabular}{llcccc}
\toprule
Task        & Description & Rank & $\alpha$ & Total $\|\Delta\|_F$ & Single-adapter score \\
\midrule
math        & MATH fine-tune (kai-xu)         & 8  & 8   & 16.62 & MATH-500 $0.164$ \\
code        & FlowerTune-Code (yangao381)     & 32 & 64  & 1.73  & HumanEval $0.598$ \\
general\_nlp & FlowerTune-NLP (zjudai)        & 32 & 64  & 1.07  & TruthfulQA-MC2 $0.557$ \\
medical     & FlowerTune-Med (yangao381)      & 32 & 64  & 1.55  & GSM8K (proxy) $0.723$ \\
\bottomrule
\end{tabular}
\end{table}

\textbf{Excluded adapters.} The two adapters flagged by the audit and
\emph{not used} in any of the 4-adapter merge experiments reported in
this paper. Both are publicly available on HuggingFace Hub, listed as
Llama-3.1-8B-Instruct adapters.

\begin{itemize}\itemsep1pt
\item \textbf{Adapter with vacuous delta.} An adapter listed for
  Llama-3.1-8B-Instruct as a ``code'' specialist has a total reconstructed delta norm $0.02$. Every forward pass is within
  floating-point noise of the base model, and task-level accuracies
  change by less than $0.5$pp on any benchmark we tested. Any
  experiment consuming this adapter would effectively run a
  $3$-adapter merge regardless of how many adapters are nominally
  included.
\item \textbf{Adapter with wrong base model.} An adapter published
  on the HuggingFace Hub under a Llama-3.1 listing has, in its
  \texttt{adapter\_config.json}, the field
  \texttt{base\_model\_name\_or\_path} pointing to
  \texttt{meta-llama/Meta-Llama-3-8B-Instruct} -- i.e.\ the config
  itself records the Llama-3 base, while the user-facing hub page
  describes the adapter as Llama-3.1-compatible. Our base-model
  match check is exactly this string comparison against the audit
  target ($=$ \texttt{NousResearch/Meta-Llama-3.1-8B-Instruct}); no
  separate detector is needed. The mismatched adapter has
  reconstructed delta norm $\sim$46. When applied to Llama-3.1, its
  delta directions are well-aligned with Llama-3 weights but largely
  uncorrelated with Llama-3.1's, so the resulting model exhibits
  large, noise-like behavioral change. Including this adapter in the
  merge produces a full-test GSM8K damage reading of $-21.7$pp;
  replacing it with an audited alternative reduces this to $-9.2$pp
  (controlled ablation in App.~\ref{app:ablation}), a $12.4$pp
  difference attributable to base-model mismatch, not merge-method behavior.
\end{itemize}

\textbf{Adapter sources and fingerprints.}
Table~\ref{tab:adapter-fingerprints} gives the HuggingFace identifier
of every candidate adapter together with a SHA256 prefix and file size
of its \texttt{adapter\_model.safetensors} (or \texttt{.bin})
checkpoint. The fingerprints pin the exact files we audited and merged,
so the results reproduce against the same checkpoints even if a
repository is later updated.

\begin{table}[h]
\centering
\scriptsize
\caption{The six Llama-3.1 candidate adapters (four audited, two
excluded) with their HuggingFace identifiers. \texttt{sha256[:16]} is
the first 16 hex digits of \texttt{sha256sum} on the actual weights
file, pinning the exact checkpoint we used.}
\label{tab:adapter-fingerprints}
\begin{tabular}{lllr}
\toprule
Slot & HuggingFace identifier & sha256 (first 16) & size \\
\midrule
\multicolumn{4}{l}{\textit{Audited (used in merge)}} \\
math         & \texttt{kai-xu/Llama-3.1-8B-Instruct-MATH-Finetuned-LoRA}        & \texttt{282fb3324fe8dd6e} & $13.0$\,MB \\
code         & \texttt{yangao381/FlowerTune-Code-Llama-3.1-8B-Instruct-PEFT}    & \texttt{44cba627372a7400} & $105.0$\,MB \\
general\_nlp & \texttt{zjudai/flowertune-general-nlp-lora-llama-3.1-8b-instruct} & \texttt{355d9b0d9469f652} & $52.0$\,MB \\
medical      & \texttt{yangao381/FlowerTune-Medical-Llama-3.1-8B-Instruct-PEFT} & \texttt{238d48683a228306} & $105.0$\,MB \\
\midrule
\multicolumn{4}{l}{\textit{Excluded by audit}} \\
no-op code      & \texttt{TianJun1/llama3.1-8b-code-reflector-lora} & \texttt{61ce8bc065cec30a} & $160.1$\,MB \\
wrong-base      & \texttt{Blackroot/Llama-3-8B-Abomination-LORA}    & \texttt{eba742709d1858b9} & $640.1$\,MB \\
\bottomrule
\end{tabular}
\end{table}

\textbf{Cross-family audit (Mistral-7B-Instruct-v0.3).} As a sanity
check we ran the same audit procedure on a candidate four-adapter set
for Mistral-7B-Instruct-v0.3 drawn from the same public hub: code
\texttt{parsak/mistralcode-7b-instruct-lora-adapters}, math
\texttt{xummer/mistral-7b-gsm8k-lora-en}, general\_nlp
\texttt{mkenfenheuer/Mistral-7B-Instruct-v0.3-ha-function-calling-lora},
and creative writing
\texttt{svjack/Mistral7B\_v2\_inst\_sharegpt\_roleplay\_chat\_lora\_small}.
Two of the four adapters fail the base-model check on minor-version
mismatch; both have non-trivial Frobenius norms and would pass a
norm-only check (Table~\ref{tab:mistral-audit}). The two adapters that
pass the audit (math, general\_nlp) are then carried into a
merging-pattern check (Table~\ref{tab:mistral-merging}), to test
whether the Llama-3.1 pattern (TA/TIES near baseline, naive drops)
reproduces on a different base.

\begin{table}[h]
\centering
\small
\caption{Cross-family audit on a candidate Mistral-7B-Instruct-v0.3
adapter set. ``Declared base'' is the value read from the adapter's
\texttt{adapter\_config.json}; the audit target is
\texttt{mistralai/Mistral-7B-Instruct-v0.3}. Both failing adapters
have non-trivial Frobenius norms; the version check is what catches
them.}
\label{tab:mistral-audit}
\begin{tabular}{llcccc}
\toprule
Task            & Declared base                            & Rank & $\alpha$ & Total $\|\Delta\|_F$ & Verdict \\
\midrule
code            & Mistral-7B-Instruct-\textbf{v0.1} & 16 & 32 & 3.625 & FAIL (version) \\
math            & Mistral-7B-Instruct-v0.3                  & 16 & 32 & 5.200 & PASS \\
general\_nlp    & Mistral-7B-Instruct-v0.3                  & 8  & 8  & 7.522 & PASS \\
creative\_writing & Mistral-7B-Instruct-\textbf{v0.2} & 8  & 16 & 7.164 & FAIL (version) \\
\bottomrule
\end{tabular}
\end{table}

\textbf{Cross-family merging (audited 2-adapter Mistral-7B-Instruct-v0.3).}
We restrict the Mistral merge experiment to the two adapters that pass
the audit (math and general\_nlp), and we report results on GSM8K only.
HumanEval pass@1 in the lm-evaluation-harness raw-completion mode
produced anomalously low scores for Mistral-7B-Instruct-v0.3 baseline
($0.043$, far below this model's published ${\sim}30$\% pass@1) while
producing the published score for Llama-3.1-8B-Instruct ($0.628$);
this is a known protocol-sensitivity of the Mistral instruct family
to chat-template handling, not a failure of the merge methods.
Re-running with proper chat-templated HumanEval is future work and
would not change the GSM8K cross-family conclusion below. Single-adapter
reference evaluations on Mistral show that, unlike Llama-3.1, the math
adapter actually \emph{helps} the protected task in isolation
($+2.9$pp over the Mistral baseline), while general\_nlp damages it
($-13.3$pp). The merge methods (Table~\ref{tab:mistral-merging})
reproduce the Llama-3.1 pattern at the magnitude family: TA
$\lambda{=}0.5$ sits within $0.6$pp of the unmerged baseline; naive
merge drops measurably; DARE $d{=}0.5$ is the worst. TIES sits
between TA and naive on Mistral, slightly below baseline, where on
Llama TIES landed nominally above baseline. We read the cross-family
check as a positive replication of the central qualitative pattern
(uniform-scaling Task Arithmetic is the most reliable simple merger),
with the quantitative point that adapter-quality regimes can vary
across model families: the same adapter task identifier need not have
the same single-adapter behavior across base models, so audits and
single-adapter references should be repeated per family.

\begin{table}[h]
\centering
\small
\caption{Cross-family merging on Mistral-7B-Instruct-v0.3 with the two
audited adapters (math, general\_nlp). Scores on full GSM8K test
(1{,}319 problems, strict-match), $\pm$ binomial stderr. $\Delta$ vs.\
Mistral baseline. Llama-3.1-8B numbers are the corresponding 4-adapter
merges from Table~\ref{tab:main} for comparison. HumanEval is omitted
for Mistral due to protocol-sensitivity (see prose).}
\label{tab:mistral-merging}
\begin{tabular}{lcc|cc}
\toprule
                 & \multicolumn{2}{c}{\textbf{Mistral-7B-v0.3}} & \multicolumn{2}{c}{\textbf{Llama-3.1-8B}} \\
Method           & GSM8K $\pm$ stderr & $\Delta$ & GSM8K & $\Delta$ \\
\midrule
Baseline (no merge)            & 0.498 $\pm$ 0.014 & ---       & 0.707 & ---       \\
\midrule
\multicolumn{5}{l}{\textit{Single-adapter reference}} \\
math alone                     & 0.527 $\pm$ 0.014 & $+0.029$  & 0.660 & $-0.047$  \\
general\_nlp alone             & 0.365 $\pm$ 0.013 & $-0.133$  & 0.654 & $-0.053$  \\
\midrule
\multicolumn{5}{l}{\textit{Merge methods (math + general\_nlp on Mistral; 4 adapters on Llama)}} \\
Naive merge                    & 0.447 $\pm$ 0.014 & $-0.051$  & 0.603 & $-0.105$  \\
Task Arith.~$\lambda{=}0.5$    & \textbf{0.492 $\pm$ 0.014} & $-0.006$  & \textbf{0.705} & $-0.002$  \\
TIES $d{=}0.5$                 & 0.465 $\pm$ 0.014 & $-0.033$  & \textbf{0.712} & $+0.005$  \\
DARE $d{=}0.5$                 & 0.419 $\pm$ 0.014 & $-0.079$  & 0.622 & $-0.085$  \\
\bottomrule
\end{tabular}
\end{table}

%============================================================
\section{Single-Adapter Reference Evaluations}
\label{app:upperbounds}
%============================================================

For each audited adapter, we evaluate the base model with only that
adapter applied (no merging, no other adapters). This decouples
adapter quality from merge quality. Table~\ref{tab:upperbounds}
reports full test-set scores.

\begin{table}[h]
\centering
\small
\caption{Single-adapter reference evaluations. $\Delta$ is the signed change
vs.\ the unmerged baseline on the same benchmark.}
\label{tab:upperbounds}
\begin{tabular}{llcc}
\toprule
Adapter     & Benchmark      & Single-adapter score & $\Delta$ vs.\ baseline \\
\midrule
Baseline (no adapter) & GSM8K        & 0.707 & --- \\
Baseline (no adapter) & HumanEval    & 0.628 & --- \\
Baseline (no adapter) & MATH-500     & 0.154 & --- \\
\midrule
math        & MATH-500       & \textbf{0.164} & $+0.010$ \\
math        & GSM8K          & 0.660 & $-0.047$ \\
code        & HumanEval      & 0.598 & $-0.030$ \\
general\_nlp & TruthfulQA-MC2 & 0.557 & $+0.026$ (vs.\ $0.531$) \\
general\_nlp & GSM8K          & 0.654 & $-0.053$ \\
medical     & GSM8K          & 0.723 & $+0.016$ (within noise) \\
\bottomrule
\end{tabular}
\end{table}

\textbf{Two observations.} (1) Only math clearly \emph{helps} its
target benchmark (MATH-500, $+1$pp); general\_nlp's TruthfulQA-MC2
gain ($+2.6$pp, ${\sim}1.7\sigma$) is borderline; code and medical are
within noise of their targets. (2) Two adapters actively
\emph{hurt} GSM8K in isolation (math $-5$pp, general\_nlp $-5$pp),
while code and medical are neutral. \emph{No single adapter helps
GSM8K.} This does not prove that adapter combinations cannot improve
GSM8K, but it means that a merge preserving the unmerged baseline
($0.707$) is already matching the strongest isolated-adapter reference
observed for the protected task.

%============================================================
\section{Negation Finding: Full Details}
\label{app:negation}
%============================================================

\begin{figure}[h]
\centering
\includegraphics[width=0.6\linewidth]{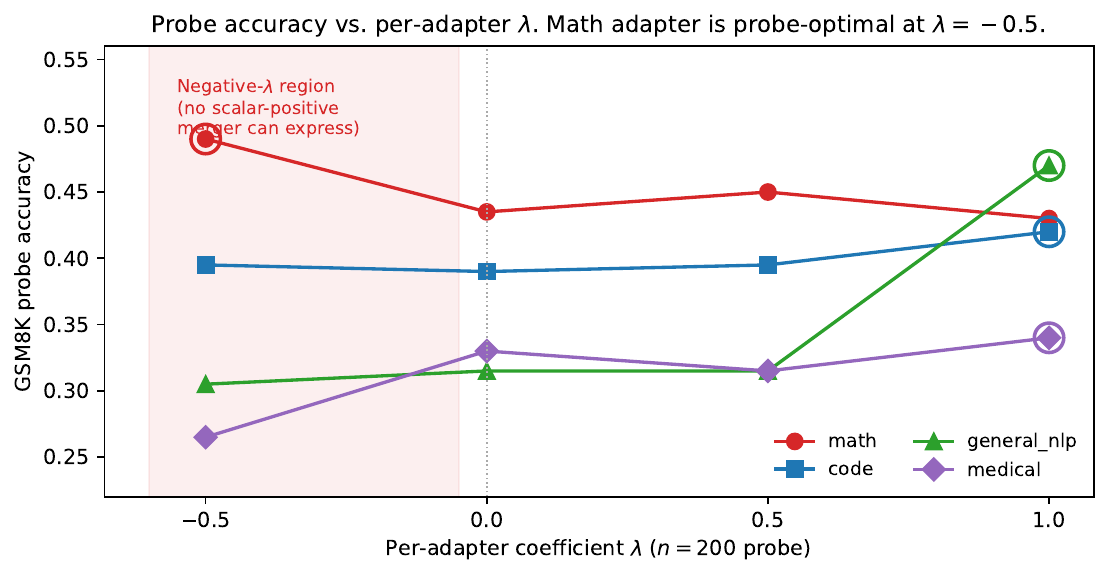}
\caption{GSM8K probe accuracy vs.\ per-adapter coefficient $\lambda$
on the audited 4-adapter Llama-3.1-8B-Instruct merge from a
gradient-free coordinate search ($n{=}200$ probe). Three adapters peak
at $\lambda{=}+1$; math peaks at $\lambda{=}-0.5$ (red region), outside
the positive-only coefficient space of TA/TIES/DARE/MagMax
(Fig.~\ref{fig:methods}).}
\label{fig:negation}
\end{figure}

Gradient-free coordinate search with candidates
$\lambda \in \{-0.5, 0, 0.5, 1.0\}$ per adapter, optimised on the
GSM8K probe set. Search trajectory (probe-set accuracy after each
candidate):

\begin{table}[h]
\centering
\small
\caption{Per-adapter probe-accuracy trajectory. Each cell is the
200-example probe accuracy after setting the named adapter's coefficient
to the column value, holding other adapters at their current value.
``Best'' is the $\lambda$ retained for that adapter before moving on:
the search adopts a candidate only if it improves the running best,
otherwise the adapter keeps its default $\lambda{=}1.0$. For
\texttt{general\_nlp} and \texttt{medical} no candidate beat the
running best $0.490$ established by \texttt{math} at $\lambda{=}-0.5$
(the per-row max for those two rows is $0.470$ and $0.340$
respectively), so $\lambda{=}1.0$ is retained as the default; the
final-configuration probe accuracy ($0.490$) coincides with the
\texttt{math} row's chosen value.}
\label{tab:negation}
\begin{tabular}{lcccc|c}
\toprule
Adapter      & $\lambda{=}-0.5$ & $\lambda{=}0$ & $\lambda{=}0.5$ & $\lambda{=}1.0$ & Best \\
\midrule
code         & 0.395 & 0.390 & 0.395 & \textbf{0.420} & $+1.0$ \\
math         & \textbf{0.490} & 0.435 & 0.450 & 0.430 & $\boldsymbol{-0.5}$ \\
general\_nlp & 0.305 & 0.315 & 0.315 & \textbf{0.470} & $+1.0$ \\
medical      & 0.265 & 0.330 & 0.315 & \textbf{0.340} & $+1.0$ \\
\bottomrule
\end{tabular}
\end{table}

The selected configuration is $\{\lambda_{\text{code}}{=}1,
\lambda_{\text{math}}{=}-0.5, \lambda_{\text{gen}}{=}1,
\lambda_{\text{med}}{=}1\}$. Probe accuracy rises from $0.405$
(naive merge) to $0.490$ (final), $+8.5$pp. On full test sets this
configuration scores GSM8K $0.672 \pm 0.013$ ($-3.5$pp from baseline
$0.707$, but $+6.9$pp above naive merge $0.603$), HumanEval
$0.555 \pm 0.039$ (within stderr of TA $\lambda{=}0.5$ at $0.561$,
$+11.0$pp above naive), and MATH-500 $0.150 \pm 0.016$ ($-0.4$pp; plasticity lost by
design since negating math removes its single-adapter MATH-500
contribution). The negation configuration is therefore strictly
better than naive merge on every benchmark, but does not beat
TA/TIES on GSM8K or MATH-500.

The \textbf{key take-away} is that the probe-optimal
$\lambda_{\text{math}}$ is \emph{negative}. No positive-coefficient
merger can access this region. A principled negation-aware merger
that preserves plasticity on tasks other than the probed one is
future work.

\begin{figure}[h]
\centering
\includegraphics[width=1\linewidth]{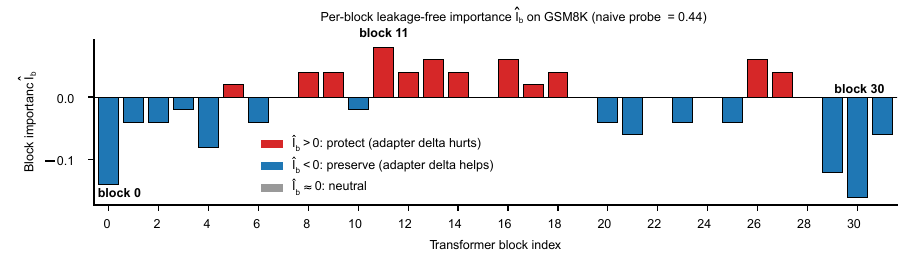}
\caption{Per-block leakage-free importance (GSM8K-train probe, $n{=}50$).
Bars are the change in probe accuracy from freezing each block's deltas
during the merge; negative bars mark blocks whose deltas help the
protected task. $14$ of $32$ blocks are negative, concentrated at the
depth boundaries.}
\label{fig:block-importance}
\end{figure}

\textbf{Structural basis from the block-importance probe.} The
negation result is not an isolated coincidence of the math adapter; it
reflects the block-level structure of the merge. The leakage-free
block-importance probe (App.~\ref{app:probing},
Fig.~\ref{fig:block-importance}) assigns each transformer block the
change in probe accuracy when that block's deltas are frozen. Of the
$32$ blocks, $14$ ($44\%$) have \emph{negative} importance: freezing
them \emph{lowers} GSM8K, i.e.\ their merged deltas are net-beneficial
for the protected task, while $12$ are positive (capability-bearing)
and $6$ are within probe noise. The negative blocks cluster at the
depth boundaries (early blocks $0$--$4$ and late blocks $29$--$31$;
the strongest, block $30$, costs $16$pp on the probe when frozen),
while the positive blocks concentrate in the middle third (blocks
$10$--$19$, mean importance $+0.03$). Positive-coefficient mergers
apply one scalar to all of a delta's directions and so cannot exploit
this sign structure; that more than $40\%$ of blocks are
net-beneficial is the structural reason a per-adapter negative
coefficient helps, and it points toward block- or direction-level
(rather than scalar-per-adapter) negation as the principled
generalisation.

%============================================================
\section{Full Results Table with Standard Errors}
\label{app:fullresults}
%============================================================

\begin{table}[h]
\centering
\small
\caption{All methods on full test sets with standard errors. Bold:
best non-baseline per column. Lower section: hyperparameter
sensitivity at $\lambda{=}0.3$ / $d{=}0.3$ for the magnitude family.}
\label{tab:full}
\begin{tabular}{lccc}
\toprule
Method                                   & GSM8K                & HE\,p@1              & MATH-500             \\
\midrule
Baseline (no merge)                       & 0.707 $\pm$ 0.013  & 0.628 $\pm$ 0.038  & 0.154 $\pm$ 0.016  \\
\midrule
Naive merge                               & 0.603 $\pm$ 0.013  & 0.445 $\pm$ 0.039  & 0.142 $\pm$ 0.016  \\
Task Arith.~$\lambda{=}0.5$               & 0.705 $\pm$ 0.013  & 0.561 $\pm$ 0.039  & 0.170 $\pm$ 0.017  \\
TIES $d{=}0.5$                            & \textbf{0.712 $\pm$ 0.012} & 0.543 $\pm$ 0.039  & 0.172 $\pm$ 0.017  \\
DARE $d{=}0.5$                            & 0.622 $\pm$ 0.013  & 0.457 $\pm$ 0.039  & 0.144 $\pm$ 0.016  \\
MagMax                                    & 0.685 $\pm$ 0.013  & 0.524 $\pm$ 0.039  & 0.168 $\pm$ 0.017  \\
\midrule
Negation finding (App.~\ref{app:negation}) & 0.672 $\pm$ 0.013 & 0.555 $\pm$ 0.039 & 0.150 $\pm$ 0.016 \\
\midrule
\multicolumn{4}{l}{\emph{Hyperparameter sensitivity (sweep)}:} \\
Task Arith.~$\lambda{=}0.3$               & 0.675 $\pm$ 0.013  & \textbf{0.598 $\pm$ 0.038}  & 0.174 $\pm$ 0.017  \\
TIES $d{=}0.3$                            & 0.708 $\pm$ 0.013  & 0.579 $\pm$ 0.039  & \textbf{0.182 $\pm$ 0.017} \\
DARE $d{=}0.3$                            & 0.603 $\pm$ 0.013  & 0.463 $\pm$ 0.039  & 0.136 $\pm$ 0.015  \\
\bottomrule
\end{tabular}
\end{table}

The lower section reports the magnitude family at $d{=}0.3$
($\lambda{=}0.3$ for TA). The qualitative ordering (TIES/TA $>$
MagMax $>$ DARE $\approx$ naive) holds at both densities; TIES at
$d{=}0.3$ ($0.708$) remains within stderr of baseline GSM8K, and
DARE remains the worst.

%============================================================
\section{Leakage-Free Probing Algorithm}
\label{app:probing}
%============================================================

\begin{algorithm}[h]
\caption{Leakage-free block importance probing.}
\label{alg:probing}
\begin{algorithmic}
\REQUIRE Base model $\theta_0$, adapters $\{\Delta_t\}_{t=1}^{T}$,
         protected task $\tau^\star$ with train set $\mathcal{D}$,
         probe size $n_{\text{probe}}$, selection size $n_{\text{sel}}$.
\STATE Split $\mathcal{D}$: probe set $P = \mathcal{D}[0{:}n_{\text{probe}}]$,
       selection set $S = \mathcal{D}[n_{\text{probe}}{:}n_{\text{probe}}+n_{\text{sel}}]$.
\STATE Evaluate $\theta_{\text{naive}} \leftarrow \theta_0 + \sum_t \Delta_t$
       on $P$; record $a_{\text{naive}} = \mathrm{acc}_P(\theta_{\text{naive}})$.
\FOR{each transformer block $b \in \{1,\dots,B\}$}
  \STATE Let $\Delta^{(-b)}_t$ = $\Delta_t$ with block-$b$ layers zeroed.
  \STATE $\theta_b' \leftarrow \theta_0 + \sum_t \Delta^{(-b)}_t$
  \STATE $\hat{I}_b \leftarrow \mathrm{acc}_P(\theta_b') - a_{\text{naive}}$
\ENDFOR
\STATE \textit{$\alpha$-selection} (if method-dependent hyperparameter):
       evaluate the merger with each candidate on $S$, pick the best.
\STATE \textbf{Test-set evaluation is run only once, after all probing
       and selection are complete.}
\end{algorithmic}
\end{algorithm}

\textbf{Disjointness.} $P$ and $S$ are disjoint by construction, and
neither intersects the held-out test set. Contrast with prior work
that uses (portions of) the test set for tuning.

\textbf{Probe set size.} We use $n_{\text{probe}}{=}50$ and
$n_{\text{sel}}{=}200$ GSM8K training examples. Compute cost at
8B-parameter scale on a single A6000: $\sim{}1.5$ hours total
(32 blocks $\times$ $\sim{}3$ min per probe eval).

%============================================================
\section{Reproducibility Details}
\label{app:repro}
%============================================================

\paragraph{Decoding parameters (all benchmarks).} All test-set
evaluations use the lm-evaluation-harness defaults for each task with
greedy decoding ($\mathrm{do\_sample}{=}\mathrm{False}$, no
temperature/top-$p$ sampling) and a single forward pass per problem.
We do not modify any of the harness's prompt formats. The relevant
generation settings:

\begin{itemize}\itemsep0pt
\item \textbf{GSM8K.} $\mathrm{max\_new\_tokens}{=}256$,
  stop tokens $\{\textnormal{``Question:''},
  \textnormal{``</s>''}, \textnormal{``<|im\_end|>''}\}$,
  exact-match (strict) scoring.
\item \textbf{HumanEval.} $\mathrm{max\_gen\_toks}{=}1024$, stop tokens
  $\{\textnormal{``$\backslash$nclass''}, \textnormal{``$\backslash$ndef''},
  \textnormal{``$\backslash$n\#''}, \textnormal{``$\backslash$nif''},
  \textnormal{``$\backslash$nprint''}\}$, pass@$1$ via the
  $\mathrm{create\_test}$ harness setting (single-sample at
  temperature 0; not 10-sample bootstrap).
\item \textbf{MATH-500.} $\mathrm{num\_fewshot}{=}4$, exact-match
  scoring on the $\boxed{\cdot}$-extracted answer; same stop tokens
  as the harness default for hendrycks\_math500.
\end{itemize}

The main tables use a single fixed seed (lm-eval default; numpy seed
$1234$, torch seed $1234$, fewshot seed $1234$). Seed and sampling
robustness are characterised in App.~\ref{app:robustness}: across three
seeds the seed-standard-deviation is $0.003$--$0.015$ on GSM8K and $0$
on greedy HumanEval, and the naive-drops / uniform-scaling-recovers
pattern persists under temperature-sampled self-consistency and
HumanEval pass@k.

\paragraph{Probe-set sizes used per analysis.} Probe sizes vary by
analysis to keep wall-clock cost tractable. Table~\ref{tab:probe-sizes}
documents each.

\begin{table}[h]
\centering
\small
\caption{Probe and selection split sizes used per experiment. All
probe and selection draws are from the GSM8K \emph{train} split,
disjoint from the GSM8K test set (1{,}319 problems) used only for
final reporting.}
\label{tab:probe-sizes}
\begin{tabular}{lll}
\toprule
Analysis & Probe size & $\alpha$-selection size \\
\midrule
Leakage-free block importance probe          & $50$ (train$[0{:}50]$) & $200$ (train$[50{:}250]$) \\
Negation finding (Table~\ref{tab:negation})  & $200$                  & --- \\
\bottomrule
\end{tabular}
\end{table}

The $n{=}200$ used in the negation coordinate search has expected
stderr ${\approx}\pm 0.035$ on binary outcomes; the four-adapter
$\times$ four-candidate sweep is run sequentially with running-best
selection. We read the finding (probe-optimal
$\lambda_{\text{math}}{=}-0.5$) as identifying an open direction
(negation-aware merging) rather than a method-level claim.

\paragraph{Compute environment.} Single NVIDIA A6000 (49 GB) for
generation; bf16 model weights; no quantisation.

%============================================================
\section{Audit Threshold Ablation and Per-Layer Norms}
\label{app:audit-ablation}
%============================================================

\textbf{Threshold sensitivity.} Our audit accepts adapters with total
delta norm $\geq 1.0$. Table~\ref{tab:audit-thresholds} shows the
admit/reject decisions over the candidate adapter set as the threshold
sweeps from $0.1$ to $5.0$. The decision is robust: any threshold in
$[0.1, 1.5]$ accepts the same four audited adapters and rejects the
norm-$0.02$ no-op. The threshold value is therefore not load-bearing
for the central audit conclusion; the qualitative gap between the
audit-failing adapter (norm $0.02$) and the lowest-norm audited
adapter (norm $1.07$) is roughly $50\times$, which is an order of
magnitude larger than any reasonable threshold uncertainty.

\begin{table}[h]
\centering
\small
\caption{Audit threshold sensitivity. ``Accept'' if total delta norm
$\geq$ threshold. Audited (used in paper) adapters are listed first;
the no-op adapter is listed last for contrast. The wrong-base-model
adapter is rejected by the base-model-match check independent of the
norm threshold.}
\label{tab:audit-thresholds}
\begin{tabular}{lc|cccccc}
\toprule
Adapter & Norm & $\geq 0.1$ & $\geq 0.5$ & $\geq 1.0$ & $\geq 1.5$ & $\geq 2.0$ & $\geq 5.0$ \\
\midrule
math (kai-xu)             & 16.62 & \checkmark & \checkmark & \checkmark & \checkmark & \checkmark & \checkmark \\
code (FlowerTune)         & 1.73  & \checkmark & \checkmark & \checkmark & \checkmark & --- & --- \\
medical (FlowerTune)      & 1.55  & \checkmark & \checkmark & \checkmark & \checkmark & --- & --- \\
general\_nlp (zjudai)      & 1.07  & \checkmark & \checkmark & \checkmark & --- & --- & --- \\
\midrule
no-op ``code'' (excluded)  & 0.02  & --- & --- & --- & --- & --- & --- \\
\bottomrule
\end{tabular}
\end{table}

\textbf{Per-layer norm distribution.} Beyond total Frobenius norm we
also examined per-layer norms for each audited adapter. The four
audited adapters show order-of-magnitude variation across layers, as
expected for low-rank PEFT modules, but no layer of any audited
adapter has norm below $10^{-3}$, i.e., none of their layers are
individually no-op. The excluded norm-$0.02$ adapter has every layer
norm below $10^{-3}$, consistent with global vanishing. We report
total norm in the main paper because (a) it is a single number that
captures the overall scale, (b) layer-wise pathologies are caught at
the global threshold for the candidates we encountered, and (c)
extending to layer-distribution audits is straightforward future
work.

%============================================================
\section{Controlled Excluded-Adapter Ablation}
\label{app:ablation}
%============================================================

To substantiate the $\sim$10pp inflation claim with attribution, we
run five naive sequential merges on the full GSM8K (1{,}319 problems)
and full HumanEval (164 problems) test sets, with the audited
4-adapter set as the reference and each excluded adapter added (or
swapped) in turn. Greedy decoding; single seed (1234).

\begin{table}[h]
\centering
\small
\caption{Excluded-adapter ablation. ``Audited 4'' is the headline
configuration of Table~\ref{tab:main}. The two excluded adapters from
App.~\ref{app:provenance} are each added on top of the audited set;
``pre-audit swap'' replaces the audited code and medical adapters with
the two excluded ones, reproducing the original (pre-audit)
candidate-pool configuration on the full test sets.
$\Delta$ is vs the unmerged baseline (GSM8K $0.707$ / HE $0.628$);
$\Delta_4$ is vs the audited-4 naive row \emph{in this table}
($0.615$ / $0.476$). The audited-4 row of Table~\ref{tab:main} reads
$0.603$ / $0.445$; the $\sim$1pp gap is the harness-version drift
documented in the ``reproduction caveat'' below.}
\label{tab:ablation}
\begin{tabular}{lcccc}
\toprule
Config & GSM8K & $\Delta$/$\Delta_4$ & HE\,p@1 & $\Delta$/$\Delta_4$ \\
\midrule
Audited 4 (this reproduction)        & $0.615$ & $-9.2$ / $\phantom{-}0.0$  & $0.476$ & $-15.2$ / $\phantom{-}0.0$  \\
Audited 4 + defective ``code''       & $0.612$ & $-9.6$ / $-0.3$            & $0.476$ & $-15.2$ / $\phantom{-}0.0$ \\
Audited 4 + wrong-base (Blackroot)   & $0.381$ & $\mathbf{-32.7}$ / $\mathbf{-23.4}$ & $0.409$ & $-21.9$ / $-6.7$ \\
Audited 4 + both excluded            & $0.367$ & $-34.0$ / $-24.8$          & $0.427$ & $-20.1$ / $-4.9$ \\
Pre-audit swap (4-adapter, original) & $\mathbf{0.491}$ & $\mathbf{-21.7}$ / $\mathbf{-12.4}$ & $0.402$ & $-22.6$ / $-7.4$ \\
\bottomrule
\end{tabular}
\end{table}

\textbf{Attribution.} Adding the defective ``code'' adapter (TianJun1,
total reconstructed norm $0.016$) to the audited 4 yields $-0.3$pp
incremental GSM8K damage --- effectively a no-op, exactly as its
near-zero norm predicts and as the independent behavioral test
confirms (App.~\ref{app:augmented}: KL$=-0.001$, top-1 agreement
$1.00$). Adding the wrong-base adapter (Blackroot,
declared base Llama-3, total reconstructed norm $46.1$) yields
$-22.2$pp incremental GSM8K damage --- it drives essentially all of
the audit-vs-no-audit gap. Adding both is no worse than adding
Blackroot alone, confirming the two excluded adapters do not interact.

\textbf{Pre-audit full-test reproduction.} The pre-audit-swap row
recovers the original (pre-audit) 4-adapter configuration on the full
test sets, giving $-21.7$pp on GSM8K (the paper's original
``$\sim$$-21$pp'' figure, previously computed on an $n{=}200$ subset,
is hereby validated on the full $n{=}1{,}319$ test set). The
audit-effect within this reproduction is therefore $21.7 - 9.2 = 12.4$pp
(the $-\Delta_4$ of the pre-audit-swap row), of which essentially all
is attributable to the wrong-base adapter alone.

\textbf{Reproduction caveat.} Our audited-4 reproduction gives GSM8K
$0.615$ vs Table~\ref{tab:main}'s $0.603$ ($+1.2$pp, within stderr
$\pm 0.013$). The qualitative pattern --- audit-effect
$\sim$11--12pp, Blackroot-attributable --- is robust to this
$\pm 1$pp drift, which we attribute to lm-evaluation-harness version
updates since the original runs.

%============================================================
\section{Prevalence Audit on a Random Sample of Public Adapters}
\label{app:prevalence}
%============================================================

The audited and excluded adapters listed in App.~\ref{app:provenance}
were hand-picked for inclusion in this study; their failure ratio is
therefore not informative as a prevalence estimate. To produce a
proper prevalence estimate, we query the HuggingFace Hub
(\texttt{api.list\_models(search=...)}) with a small set of
\texttt{lora <base>} variants per family, deduplicate across queries,
exclude any adapter that already appears in our curated pool
(to avoid double-counting our own choices), and random-shuffle the
remainder with a fixed seed (1234) before taking the first $N{=}40$
per family for audit.

\begin{table}[h]
\centering
\small
\caption{Random-sample prevalence audit, $N{=}40$ per base-model family.
``Loadable'' means \texttt{adapter\_config.json} and weights file both
loaded. Rates ``of loadable'' are computed conditional on loadability.}
\label{tab:prevalence}
\begin{tabular}{lcccc}
\toprule
Family & Loadable & Pass audit & Base mismatch (of loadable) & Near-zero norm (of loadable) \\
\midrule
Llama-3.1-8B-Instruct   & $35/40$ ($87.5\%$) & $6/40$ ($15.0\%$) & $15/35$ ($42.9\%$) & $10/35$ ($28.6\%$) \\
Mistral-7B-Instruct-v0.3 & $27/40$ ($67.5\%$) & $0/40$ ($\mathbf{0.0\%}$) & $26/27$ ($\mathbf{96.3\%}$) & $1/27$ ($3.7\%$) \\
\midrule
Combined ($N{=}80$)     & $62/80$ ($77.5\%$) & $6/80$ ($7.5\%$)  & $41/62$ ($\mathbf{66.1\%}$) & $11/62$ ($17.7\%$) \\
\bottomrule
\end{tabular}
\end{table}

\textbf{Findings.} Across both families, only $6/80$ random-sampled
adapters ($7.5\%$) pass the audit for their declared base. The
dominant failure mode is base mismatch ($66\%$ of loadable adapters
target a different base model than the one we specified); near-zero
norms account for an additional $18\%$. The Mistral case is
particularly stark: $0/40$ random-sampled ``mistral 7b'' LoRAs target
v0.3 by declared base; almost all target v0.1 or v0.2. The two
audited Mistral adapters in our paper are therefore real positive
outliers in this distribution, not representative draws.

\textbf{Caveats.} (i) The sample is biased by HuggingFace search's
default ranking (relevance / likes), not strictly uniform random over
all uploaded LoRAs; this is the distribution a user encounters when
picking by search, which is the operationally relevant one for the
``audit before you merge'' claim. (ii) ``Base mismatch'' includes
adapters that are perfectly fine for the base they actually target
(e.g., a v0.2 adapter is fine for v0.2 users); the prevalence we
report is of ``would not work if naively used for our declared base'',
which is precisely the misconfiguration scenario the paper addresses.
(iii) $N{=}40$ per family yields a binomial standard error
${\sim}5\%$ around the pass rate.

\textbf{Cross-scale prevalence.} To test whether these findings are
specific to the $7$--$8$B scale, we repeated the random-sample audit
($N{=}40$ each, same seed) on Llama-3.2-1B, Llama-3.2-3B, and
Qwen2.5-7B (Table~\ref{tab:prevalence-scale}). Pass rates remain
uniformly low ($0$--$15\%$) and base-model mismatch remains the
dominant failure mode ($43$--$96\%$ of loadable adapters) across every
scale and both architecture families. Near-zero-norm prevalence is
more pool-specific ($0$--$29\%$): the norm check is load-bearing for
some pools and not others, but the base-model check is load-bearing
everywhere, consistent with our design rationale that both checks are
necessary and neither is redundant. The audit's necessity is therefore
not scale- or family-specific; if anything smaller-scale and non-Llama
pools have \emph{higher} base-mismatch rates.

\begin{table}[h]
\centering
\small
\caption{Cross-scale random-sample prevalence audit ($N{=}40$ per
family, seed 1234). Rates ``of loadable'' are conditional on the
adapter loading. The $8$B and Mistral rows are from
Table~\ref{tab:prevalence}.}
\label{tab:prevalence-scale}
\begin{tabular}{llccc}
\toprule
Scale & Family & Pass audit & Base mismatch & Near-zero norm \\
\midrule
1B & Llama-3.2-1B    & $4/40$ ($10.0\%$)  & $81\%$ & $8\%$  \\
3B & Llama-3.2-3B    & $5/40$ ($12.5\%$)  & $81\%$ & $0\%$  \\
7B & Qwen2.5-7B      & $3/40$ ($7.5\%$)   & $91\%$ & $3\%$  \\
7B & Mistral-7B-v0.3 & $0/40$ ($0.0\%$)   & $96\%$ & $4\%$  \\
8B & Llama-3.1-8B    & $6/40$ ($15.0\%$)  & $43\%$ & $29\%$ \\
\bottomrule
\end{tabular}
\end{table}

%============================================================
\section{Augmented Audit: Tokenizer, Base-Signature, and Output-Divergence Signals}
\label{app:augmented}
%============================================================

The two-check audit (Sec.~\ref{sec:setup}) relies on metadata strings
and a reconstructed-delta-norm threshold. To address the concern that
adversarial or mistaken metadata could pass these checks, we
supplement them with three independent content-based signals:

\begin{itemize}\itemsep1pt
\item \textbf{Base-checkpoint hash signature.} SHA256 of the base
  model's \texttt{config.json} and first three safetensors shards.
  Lets a verifier confirm they use the same base checkpoint we
  audited. Signatures: Llama-3.1-8B-Instruct
  \texttt{e3622c38fbff0f8d}; Mistral-7B-Instruct-v0.3
  \texttt{ee38818a6b9e1d30}.
\item \textbf{Tokenizer compatibility.} For each adapter we (a) check
  whether \texttt{target\_modules} include vocab-dependent layers
  (\texttt{embed\_tokens}, \texttt{lm\_head}), and if so whether the
  lora-A/B dimensions match the base \texttt{vocab\_size}; and
  (b) hash any tokenizer files shipped with the adapter and compare
  to the base.
\item \textbf{Output-divergence sanity test.} On a fixed set of
  $10$ short prompts, we run the adapted and unadapted base in
  parallel and report (i) mean KL divergence on the next-token
  distribution, (ii) top-1 token agreement rate, and (iii)
  character-Jaccard similarity of $32$-token greedy generations.
\end{itemize}

\begin{table}[h]
\centering
\small
\caption{Augmented audit signals on the candidate set. ``Meta'' is the
two-check verdict from Sec.~\ref{sec:setup}; ``Tok'' is the tokenizer
check from this section; KL / top-1 / sim are the divergence test
outputs. ``Interp'' is a heuristic interpretation. Llama only; Mistral
follows the same pattern.}
\label{tab:augmented}
\begin{tabular}{lccccccc}
\toprule
Adapter (slot)                & Meta & Tok  & KL    & top-1 & gen-sim & Interpretation \\
\midrule
math (audited)                & PASS & ok   & $0.42$ & $0.60$ & $0.67$ & plausible effect \\
code (audited)                & PASS & ok   & $0.04$ & $0.80$ & $0.87$ & plausible effect \\
general\_nlp (audited)        & PASS & ok   & $0.01$ & $0.70$ & $0.84$ & plausible effect \\
medical (audited)             & PASS & ok   & $0.06$ & $0.60$ & $0.82$ & plausible effect \\
code-defective (TianJun1)     & FAIL & warn & $-0.00$ & $\mathbf{1.00}$ & $\mathbf{1.00}$ & \textbf{no-op (confirmed)} \\
wrong-base (Blackroot)        & FAIL & ok   & $0.10$ & $0.80$ & $0.75$ & plausible effect (!) \\
\bottomrule
\end{tabular}
\end{table}

\textbf{Per-failure-mode coverage.} The three signals are
complementary, not redundant:

\begin{itemize}\itemsep1pt
\item \emph{No-op (defective TianJun1):} caught by the metadata norm
  check and independently confirmed by the divergence test
  (KL\,$\approx 0$, top-1 agreement $1.00$). The tokenizer check also
  flags shipped tokenizer files that hash-mismatch the base.
\item \emph{Wrong base (Blackroot, Llama-3 vs Llama-3.1):} caught by
  the metadata base-model match only. The divergence test on short
  prompts does \emph{not} flag it (KL\,$=0.10$, in the same range as
  legitimate adapters), because Llama-3 and Llama-3.1 share
  architecture and most training, so the wrong-base delta still
  produces coherent next-token distributions. The damage shows up
  downstream (App.~\ref{app:ablation}: $-22.2$pp on full-test GSM8K)
  but cannot be detected with a short-prompt logit divergence.
\end{itemize}

We therefore retain the metadata base-model match as a necessary
audit check: behavioral testing on short prompts is a useful
\emph{additional} signal for near-zero adapters, not a substitute for
the metadata check on architecturally-close-but-wrong base models.  More sophisticated divergence signals --- e.g., variance-based anomaly detection \citep{aharon2026lightning} --- could catch pathologies that pass both metadata and short-prompt logit-divergence checks; we leave a systematic adaptation to future work.

%============================================================
\section{Seed and Sampling Robustness}
\label{app:robustness}
%============================================================

The main results use single-seed greedy decoding. We check robustness
along two axes: multiple random seeds, and
temperature sampling (GSM8K self-consistency and HumanEval pass@k).

\textbf{Seed robustness (greedy, 3 seeds).} Re-running each method
across seeds $\{1234, 5678, 9012\}$ gives seed-standard-deviations of
$0.003$--$0.015$ on GSM8K and exactly $0$ on HumanEval (greedy code
generation is deterministic). Seed variance is an order of magnitude
smaller than the method differences, so the single-seed ranking is
not a seed artefact: baseline $0.702{\pm}0.003$, naive
$0.604{\pm}0.010$, TA $0.698{\pm}0.010$, TIES $0.715{\pm}0.015$ on
GSM8K.

\textbf{Sampling robustness.} Table~\ref{tab:robustness} reports GSM8K
self-consistency@8 (temperature $0.7$, majority vote over $8$ samples,
$500$-question subset) and HumanEval pass@$\{1,10\}$ (temperature
$0.2$, $10$ samples). The audit-merge conclusion holds under
stochastic decoding, not just greedy: under self-consistency, TA
($0.844$) and TIES ($0.856$) sit at or within a point of the unmerged
baseline ($0.856$) while naive drops to $0.760$; under HumanEval
pass@10, TA ($0.744$) slightly \emph{exceeds} the baseline ($0.732$)
and TIES matches it, while naive lags at $0.604$. The
naive-drops / uniform-scaling-recovers pattern is therefore not an
artefact of greedy decoding.

\begin{table}[h]
\centering
\small
\caption{Sampling robustness on the audited 4-adapter set. GSM8K:
self-consistency@8 (majority vote, $T{=}0.7$) and greedy on the same
$500$-question subset (first $500$ test indices, evaluated with a
custom $4$-shot chain-of-thought prompt rather than lm-eval's
strict-match harness; the resulting baseline greedy of $0.814$ is
${\sim}11$pp above the full-test lm-eval baseline of $0.707$ in
Table~\ref{tab:main}, so absolute numbers here are not directly
comparable across tables --- the claim of this table is the
\emph{cross-method ordering}, which matches the full-test ordering).
HumanEval: pass@1 and pass@10 ($T{=}0.2$, $10$ samples, lm-eval
\texttt{repeats} mechanism).}
\label{tab:robustness}
\begin{tabular}{lcccc}
\toprule
Method & GSM8K SC@8 & GSM8K greedy & HE pass@1 & HE pass@10 \\
\midrule
Baseline           & 0.856 & 0.814 & 0.612 & 0.732 \\
Naive              & 0.760 & 0.676 & 0.482 & 0.604 \\
TA $\lambda{=}0.5$ & 0.844 & 0.792 & 0.587 & \textbf{0.744} \\
TIES $d{=}0.5$     & \textbf{0.856} & 0.782 & 0.583 & 0.732 \\
\bottomrule
\end{tabular}
\end{table}

%============================================================
\section{Per-Step Forgetting Curves}
\label{app:perstep}
%============================================================

The main results (Table~\ref{tab:main}) report scores after all four
adapters are merged. To expose the continual-learning dynamics, we
evaluate the protected task (GSM8K) and HumanEval after \emph{each}
sequential merge step on the audited 4-adapter set
(order code $\to$ math $\to$ general\_nlp $\to$ medical).

\begin{table}[h]
\centering
\small
\caption{Per-step GSM8K accuracy after each adapter merge (audited set,
full test). Unmerged baseline $0.707$. The final column comes from a later run with the current harness
(its naive entry, $0.615$, matches the reproduction in
App.~\ref{app:ablation}), so it differs from the GSM8K column of
Table~\ref{tab:main} by $+1.2$pp (naive), $-0.4$pp (TA), $+0.7$pp
(TIES), $-4.2$pp (DARE), and $+0.1$pp (MagMax).}
\label{tab:perstep}
\begin{tabular}{lcccc}
\toprule
Method & +code & +math & +nlp & +medical \\
\midrule
Naive          & 0.685 & 0.651 & 0.629 & 0.615 \\
TA $\lambda{=}0.5$ & 0.683 & 0.707 & 0.688 & 0.701 \\
TIES $d{=}0.5$ & 0.691 & 0.713 & 0.692 & 0.719 \\
DARE $d{=}0.5$ & 0.704 & 0.647 & 0.624 & 0.580 \\
MagMax         & 0.678 & 0.699 & 0.684 & 0.686 \\
\bottomrule
\end{tabular}
\end{table}

The per-step curves split the mergers into two regimes that the
aggregate table alone does not show. \textbf{Naive and DARE degrade
monotonically} ($0.685 \to 0.615$ and $0.704 \to 0.580$), with each
added adapter compounding the loss --- the signature of catastrophic
forgetting. \textbf{TA, TIES, and MagMax recover to near baseline
after the first step and stay there} for every subsequent merge
($0.68$--$0.72$ at every step for TA/TIES). The HumanEval
trajectories show the same split: naive drops to $0.476$ by the final
step while TA/TIES/MagMax hold $0.55$--$0.59$. The final-step ordering TIES $>$ TA $>$ MagMax $>$ \{naive, DARE\}
matches Table~\ref{tab:main}; only naive and DARE swap places between
this run and Table~\ref{tab:main} (DARE $0.580$ vs.\ naive $0.615$ here;
$0.622$ vs.\ $0.603$ there), so we do not rank those two against each
other. The two-regime split is therefore not an artefact of where the
trajectory happens to end.

%============================================================
\section{Orthogonal-Projection Merger Under the Audit}
\label{app:orthogonal}
%============================================================

To partially address the question of whether the audit's effect
generalises to LoRA-aware mergers, we add a controlled
\textbf{orthogonal-projection merger} inspired by OSRM. At each
sequential merge step, the new adapter's delta is projected onto the
orthogonal complement of the column space of all previously-applied
deltas before being added:
$$
  \Delta_{\text{orth}}^{(t)} \;=\; \Delta^{(t)} \;-\; U_{\text{cum}}^{(t-1)} \big(U_{\text{cum}}^{(t-1)}\big)^{\!\top} \Delta^{(t)}
$$
where $U_{\text{cum}}^{(t-1)}$ is a cumulative orthonormal basis of
the union $\mathrm{span}\{\Delta^{(1)},\dots,\Delta^{(t-1)}\}$ at each
layer, computed incrementally via thin SVD of $\Delta_{\text{orth}}$
at each step. The math is equivalent to stacking all priors and
SVD-projecting; the incremental form uses
$O(\text{out\_dim} \times \text{cumulative\_rank})$ storage rather
than $O(\text{out\_dim} \times \text{in\_dim} \times T)$.

\textbf{Disclaimer.} This is \emph{not} a faithful reproduction of
SA-LoRA~\citep{li2025salora}, OSRM~\citep{osrm2025},
LoRA-LEGO~\citep{lora_lego2024}, or LoRI~\citep{lori2025}; it is a
controlled implementation of the family's central idea
(orthogonalise before merge) using only public LoRA factorisations.
Faithful reproductions of the four named methods under our audit
protocol remain open work.

\begin{table}[h]
\centering
\small
\caption{Orthogonal-projection merger on the audited and pre-audit
4-adapter sets; comparison with naive merge from App.~\ref{app:ablation}.
Full GSM8K (1{,}319) and HumanEval (164) test sets; greedy decoding;
single seed.}
\label{tab:orthogonal}
\begin{tabular}{llcccc}
\toprule
Merger             & Pool          & GSM8K & HE & GSM8K $\Delta$ vs unmerged \\
\midrule
\multirow{2}{*}{Naive (App.~\ref{app:ablation})} & Audited       & $0.615$ & $0.476$ & $-9.2$ \\
                                                  & Pre-audit swap & $0.491$ & $0.402$ & $-21.7$ \\
\midrule
\multirow{2}{*}{Orthogonal proj.}                 & Audited       & $\mathbf{0.635}$ & $0.476$ & $-7.3$ \\
                                                  & Pre-audit swap & $\mathbf{0.523}$ & $0.476$ & $-18.4$ \\
\bottomrule
\end{tabular}
\end{table}

\textbf{Audit-effect comparison.}

\begin{center}
\begin{tabular}{lcc}
\toprule
Merger             & Audited GSM8K & Pre-audit GSM8K \\
                   & \multicolumn{2}{c}{Audit-effect (audited $-$ pre-audit)} \\
\midrule
Naive              & $0.615$  $\Leftarrow$ & $0.491$ \\
                   & \multicolumn{2}{c}{$+12.4$pp} \\
Orthogonal proj.   & $0.635$  $\Leftarrow$ & $0.523$ \\
                   & \multicolumn{2}{c}{$\mathbf{+11.1}$pp} \\
\bottomrule
\end{tabular}
\end{center}

\textbf{Key reading.} The audit improves the orthogonal merger by
$+11.1$pp on GSM8K --- within $1$pp of its effect on naive merging
($+12.4$pp). Switching from naive to orthogonal-projection merging
on the audited pool gains $+2$pp; switching on the pre-audit pool
gains $+3.3$pp. The audit's effect (${\sim}11$pp) is therefore an
order of magnitude larger than the structure-aware-vs-naive
difference (${\sim}2$--$3$pp).

\textbf{Interpretation.} Structure-aware merging is mildly
complementary to adapter auditing on this set: the orthogonal
projection partially cancels the wrong-base adapter's harmful
directions when they overlap with the already-applied audited
adapters' subspaces, recovering $3.3$pp on the pre-audit pool. But
this gain is small relative to the audit-vs-no-audit gap, and the
ordering (orthogonal $>$ naive) is preserved by the audit on both
pools. The result supports the paper's central position: in this
setting, adapter-quality controls dominate merger-method choice
on the protected task, even for a representative structure-aware
merger.

\end{document}